\pdfoutput=1
\documentclass[11pt]{article}

\usepackage[a4paper,margin=1in]{geometry}
\usepackage[T1]{fontenc}
\usepackage[utf8]{inputenc}
\usepackage{lmodern}
\usepackage{microtype}
\usepackage{amsmath,amssymb,bm}
\usepackage{booktabs}
\usepackage{multirow}
\usepackage{array}
\usepackage{graphicx}
\usepackage{xcolor}
\usepackage{tikz}
\usetikzlibrary{arrows.meta,positioning,fit,calc}
\usepackage[numbers,sort&compress]{natbib}
\usepackage[hidelinks]{hyperref}
\usepackage{caption}
\usepackage{enumitem}
\setlist{itemsep=2pt,topsep=4pt}

\newcommand{\ind}{\mathbf{1}}

\title{\textbf{RubriX: A Fast, Fine-Tunable\\Non-Autoregressive Rubric Judge}}
\author{Balamurugan Paramasivam\\ CobuildX\\ \texttt{balamurugan.p@cobuildx.ai} \and Sudalai Rajkumar\\ CobuildX\\ \texttt{srk@cobuildx.ai}}
\date{}

\begin{document}
\maketitle

\begin{abstract}
Rubric-based evaluation asks which level of a rubric a response meets. It is usually done by prompting a large language model (LLM) as a judge, which needs a generative call per judgment and reads every level description, so its cost grows with the rubric. We present RubriX, a rubric judge whose architecture is inspired by contrastive language models (CLM): a LoRA-adapted encoder embeds the input and each level description, and a softmax over scaled cosine similarities gives a probability for every level without text generation. Trained on public rubric data and proprietary synthetic data, RubriX reaches 75.4\% exact accuracy on held-out rubrics, against 57.1\% for TypeSafe's Jev and 49.8\% for a Qwen3-8B judge. On eight human-labelled benchmark sets it is within 6 percentage points of Jev's accuracy on most of them, while Jev keeps a larger lead on two and ranks responses best on average; against the results reported in the G-Eval paper, RubriX outperforms G-Eval with GPT-3.5 on every SummEval dimension and both QAGS subsets. With level embeddings cached, RubriX's latency stays at about 28\,ms per judgment as level descriptions grow from 10 to 3,000 tokens, while Jev's (approximate network time subtracted) rises from 44\,ms to 211\,ms and the Qwen3-8B judge's from 32\,ms to 516\,ms, making RubriX 7.7$\times$ faster than Jev and 18.8$\times$ faster than the Qwen3-8B judge at 3,000 tokens per level.
\end{abstract}

\section{Introduction}

Evaluating AI systems increasingly comes down to one repeated decision: given a response and a rubric, which level of the rubric does the response meet? This is usually done by prompting a large language model as a judge \citep{zheng2023judging}, as in the rubric-based metrics of Ragas \citep{es2024ragas,ragasdocs} and the G-Eval metric \citep{liu2023geval} implemented in DeepEval \citep{deepeval}. These approaches are flexible, since any criterion can be written as text, but every judgment needs a full generative LLM call over the whole rubric, which makes them slow and costly when applied to every response and to long rubrics. Their scores also depend on prompt wording, decoding settings and output parsing.

RubriX takes a different approach, with an architecture inspired by contrastive language models (CLM) \citep{clm}. The response and each rubric level are embedded with a language-model encoder, and small trained heads score how well the response matches each level. A rubric is scored without text generation, returning a probability for every level, and the level embeddings can be computed once per rubric and reused for every response. Our contributions are:
\begin{itemize}
  \item \textbf{A non-autoregressive rubric judge whose cost does not grow with the rubric.} RubriX scores a response against a rubric's own level descriptions with a bi-encoder: one LoRA-adapted encoder embeds the input and each level separately, so a rubric of any size is scored without generation, and its level embeddings are computed once and reused for every response it judges.
  \item \textbf{Ordinal training for a CLM-inspired architecture.} The original CLM treats options as unordered. We train with cross-entropy over each rubric's own levels plus a ranked probability score whose weight adapts to the running magnitudes of the two losses, so the ordinal term keeps about an equal share of the objective throughout training without a hand-tuned weight. Weighting losses adaptively from their running statistics is established in multi-task learning \citep{chen2018gradnorm,groenendijk2021cov}; we apply a simple form of it to the ordinal term.
  \item \textbf{Training data.} RubriX is trained on a criterion-capped mix of public rubric datasets and a proprietary dataset.
  \item \textbf{Evaluation and latency.} We compare RubriX with Jev \citep{jev}, a Qwen3-8B judge \citep{qwen3} and the published G-Eval results \citep{liu2023geval} on held-out rubrics and eight human-labelled benchmark sets, by accuracy and by ranking correlation, and show that with cached level embeddings its latency stays flat as rubrics gain levels and longer level descriptions, while judges that read the levels slow down.
\end{itemize}
The RubriX-8B model is open source and available on Hugging Face at \url{https://huggingface.co/CobuildX-AI/RubriX-8B}, together with code to load it and score rubrics.

\section{Related Work}

\paragraph{LLM-as-a-judge.} Strong LLMs agree well with human preferences when prompted to evaluate responses \citep{zheng2023judging}, which has made them the default evaluator for open-ended generation. They are, however, sensitive to prompt format and position, and every judgment is a generative call. A rubric judge of this kind also places the level descriptions inside its prompt, so every judgment reprocesses the whole rubric: its cost grows with the number of levels and with the length of each description, and the rubric must fit in the model's context window together with the input. In our measurements (Section~\ref{sec:latency}), a Qwen3-8B judge with a 32k context holds at most 20 levels of 1,500 tokens each.

\paragraph{Rubric-based evaluation in Ragas.} Ragas \citep{es2024ragas} provides a rubric-based scoring metric for LLM outputs \citep{ragasdocs}. The user defines a rubric as a description for each score level, typically 1 to 5. An LLM judge reads the input, the response and optionally a reference, then returns the score whose description fits best, along with written feedback. Ragas also supports instance-specific rubrics, where each example carries its own level descriptions.

\paragraph{G-Eval.} G-Eval \citep{liu2023geval} uses an LLM with chain-of-thought to evaluate generated text against user-defined criteria. The model first expands the criteria into evaluation steps, then scores the output in a form-filling step. Instead of taking the single generated score, G-Eval weights each possible score by the probability the model assigns to it, which gives finer-grained scores with better alignment to human judgments. DeepEval \citep{deepeval} implements this approach as its GEval metric for custom criteria. We compare with G-Eval's published results on SummEval and QAGS.

\paragraph{Jev.} Jev from TypeSafe \citep{jev} is a hosted model built for decisions rather than generation. Given a context and a set of options, it returns a probability for each option in a single pass, with no text to generate or parse. Rubric scoring maps onto it directly: the criterion becomes the question and the level descriptions become ordered options. Jev is available as a hosted API; its weights are not released, and a score question accepts at most 10 options.

\paragraph{Contrastive scoring.} Contrastive objectives such as InfoNCE \citep{oord2018cpc} and CLIP \citep{radford2021clip} learn a shared embedding space in which matching pairs score highly, using a temperature-scaled cosine similarity. CLM \citep{clm} applies this idea to choice tasks with a frozen LLM encoder and trained projection heads. RubriX's architecture is inspired by CLM; it adapts the encoder and trains with an ordinal objective for rubric scoring.

\paragraph{Summary.} Ragas and G-Eval treat evaluation as a generation task, with the cost and sensitivity that implies. Jev shows that scoring the rubric's options directly is practical; its weights are not released. This leaves room for a rubric scorer that scores options directly and that can also be trained and adapted openly.

\section{Method}

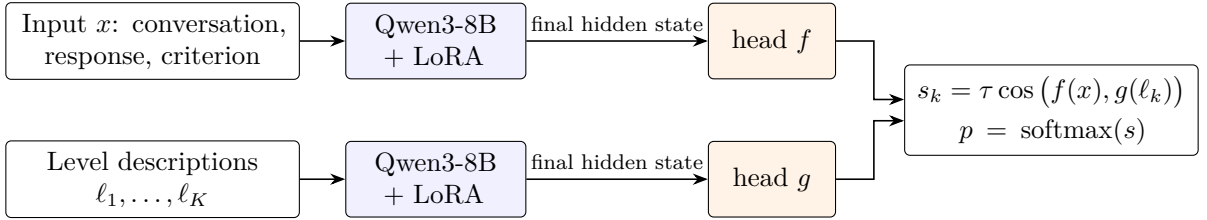
\begin{figure}[!htb]
\centering
\begin{tikzpicture}[
  box/.style={draw,rounded corners=2pt,align=center,font=\small,minimum height=1cm,inner sep=4pt},
  enc/.style={box,fill=blue!6,text width=2.1cm},
  head/.style={box,fill=orange!10,text width=1.4cm},
  arr/.style={-{Stealth[length=2mm]},semithick}]
\node[box,text width=3.6cm] (x) {Input $x$: conversation,\\response, criterion};
\node[box,text width=3.6cm,below=0.8cm of x] (l) {Level descriptions\\$\ell_1,\dots,\ell_K$};
\node[enc,right=0.6cm of x] (ex) {Qwen3-8B\\+ LoRA};
\node[enc,right=0.6cm of l] (el) {Qwen3-8B\\+ LoRA};
\node[head,right=2.4cm of ex] (f) {head $f$};
\node[head,right=2.4cm of el] (g) {head $g$};
\node[box,right=0.9cm of $(f.east)!0.5!(g.east)$,text width=3.6cm] (s) {$s_k=\tau\cos\big(f(x),g(\ell_k)\big)$\\[3pt]$p=\mathrm{softmax}(s)$};
\draw[arr] (x) -- (ex); \draw[arr] (l) -- (el);
\draw[arr] (ex) -- node[above,font=\scriptsize]{final hidden state} (f);
\draw[arr] (el) -- node[above,font=\scriptsize]{final hidden state} (g);
\draw[arr] (f.east) -- ++(0.5,0) |- ([yshift=4pt]s.west);
\draw[arr] (g.east) -- ++(0.5,0) |- ([yshift=-4pt]s.west);
\end{tikzpicture}
\caption{The RubriX scorer. The input and each level description are encoded separately by the same encoder, projected by two heads, and compared by temperature-scaled cosine similarity. A softmax over the rubric's own $K$ levels gives a probability for each level. Level embeddings do not depend on the input, so they can be computed once per rubric and cached.}
\label{fig:model}
\end{figure}

\subsection{Scoring model}
RubriX scores a response against a rubric with a contrastive approach (Figure~\ref{fig:model}). The input $x$ (the conversation, the response to evaluate and the criterion) and each rubric level description $\ell_k$ are embedded separately by a language-model encoder, Qwen3-8B \citep{qwen3}, with no text generation. Each text's embedding is its final hidden state, the last layer's hidden state at the last token, L2-normalised. The same encoder, with one shared set of adapters (Section~3.3), embeds both. Two projection heads, $f$ for inputs and $g$ for levels, each a three-layer MLP ($4096\!\to\!1536\!\to\!1536\!\to\!512$ with LayerNorm and GELU), map the embeddings into a shared space. For a rubric with $K$ levels, the score of level $k$ is
\begin{equation}
  s_k = \tau \cdot \cos\big(f(x),\, g(\ell_k)\big),
\end{equation}
and the predicted probability of level $k$ is
\begin{equation}
  p_k = \frac{\exp(s_k)}{\sum_{j=1}^{K} \exp(s_j)} .
\end{equation}
The temperature $\tau = 100$ is inherited from CLM, where it is a learned parameter capped at 100. Each example competes only against the levels of its own rubric, never against other rubrics' levels.

\subsection{Training objective}
For a gold level $y$, the primary objective is cross-entropy over the rubric's levels:
\begin{equation}
  \mathcal{L}_{\mathrm{CE}} = -\log p_y .
\end{equation}
Cross-entropy treats the levels as unordered: predicting a level two steps away costs the same as predicting a neighbouring one. We add the ranked probability score (RPS) \citep{epstein1969rps}, which compares the predicted cumulative distribution with the gold one:
\begin{equation}
  \mathcal{L}_{\mathrm{RPS}} = \frac{1}{K-1} \sum_{k=1}^{K-1} \Big( P_k - \ind[y \le k] \Big)^2,
  \qquad P_k = \sum_{j=1}^{k} p_j .
\end{equation}
Here $P_k$ is the predicted probability that the level is at most $k$. A prediction one level from the gold level costs less than one several levels away. RPS is much smaller than cross-entropy in magnitude, so a fixed weight lets it contribute little. We therefore weight it adaptively, so that on average it contributes as much as cross-entropy:
\begin{equation}
  \mathcal{L} = \mathcal{L}_{\mathrm{CE}} + \lambda_t\, \mathcal{L}_{\mathrm{RPS}} ,
  \qquad \lambda_t = \alpha\,\frac{\overline{\mathcal{L}}_{\mathrm{CE}}}{\overline{\mathcal{L}}_{\mathrm{RPS}}},
\end{equation}
where the bars are exponential moving averages (decay 0.99) of the two losses over micro-batches, computed from detached values so that no gradient flows through $\lambda_t$, seeded with the first micro-batch and capped at 1000; we use $\alpha = 1$. This is a simple instance of adaptive loss balancing \citep{chen2018gradnorm,groenendijk2021cov}: instead of a fixed weight, each term is scaled by its own running magnitude.

\subsection{Encoder adaptation}
Low-rank adapters \citep{hu2022lora} are added to all attention and MLP projections of the encoder and trained jointly with the heads, so that the encoder itself can learn features that separate adjacent levels. This trains 43.6M adapter parameters (about 0.5\% of the encoder) and 18.9M head parameters. The heads start from the released CLM-8B heads \citep{clm}. For serving, the adapters are merged into the encoder weights, so inference has the cost of the base encoder.

\section{Experimental Setup}

\subsection{Training data}
The training set (Table~\ref{tab:train}) has 34,261 examples covering 2,284 distinct criteria. It combines public rubric-labelled data with a proprietary dataset. The proprietary dataset contains 8,039 rubric-labelled examples over 324 criteria, covering long-form answers to expert questions, multi-turn conversations, and other open-ended writing tasks. It was generated with a large language model and filtered by blind re-grading: an example was kept only when an independent grader, without seeing the intended level, assigned exactly that level. The public part is drawn from seven sources and capped at 25 examples per criterion, balanced across levels within each criterion, so that sources with a handful of fixed criteria (HelpSteer's attributes, for example) contribute few examples and the many custom rubrics of the Feedback Collection \citep{kim2024prometheus} and the principles of HelpSteer3 \citep{helpsteer3} contribute most. The test set is split from training by criterion: no criterion or level description of the held-out test set appears in training, and some criteria (HelpSteer's complexity and HelpSteer2's verbosity, for example) are held out entirely. The validation data, used only to select the checkpoint, partly shares criteria with training.

\begin{table}[!htb]
\centering\footnotesize
\setlength{\tabcolsep}{4pt}
\caption{Training data. Public sources are capped at 25 examples per criterion, level-balanced.}
\label{tab:train}
\begin{tabular}{@{}lrlr@{}}
\toprule
Source & Criteria & Levels & Examples \\
\midrule
Feedback Collection \citep{kim2024prometheus} & 898 & 5 & 22,450 \\
HelpSteer3, principles \citep{helpsteer3} & 1,046 & 2 & 3,372 \\
HelpSteer, HelpSteer2 \citep{wang2024helpsteer,wang2024helpsteer2} & 8 & 5 & 200 \\
HelpSteer3, feedback \citep{helpsteer3} & 3 & 5 & 75 \\
UltraFeedback \citep{cui2024ultrafeedback} & 4 & 5 & 100 \\
ASAP 2.0 \citep{asap2} & 1 & 6 & 25 \\
Proprietary dataset & 324 & 2--5 & 8,039 \\
\midrule
Total & 2,284 & & 34,261 \\
\bottomrule
\end{tabular}

\end{table}

\subsection{Benchmarks}
We evaluate on six human-labelled benchmarks (Table~\ref{tab:bench}). JudgmentBench scores each answer with pass/fail checklist items and with penalty items that count errors; we report these separately as JudgmentBench, binary and JudgmentBench, multi-level, with the counts grouped into four levels (none, one, a few, many). QAGS is reported per subset. Rubric text is kept verbatim where the benchmark provides it. Where a benchmark has no level descriptions (HealthBench and the JudgmentBench checklist items are met/not-met, JudgmentBench's penalty items are counts, SummEval has 1--5 ratings and QAGS a yes/no question), short generic level descriptions were added, the same for every judge, since there is no other way to pose those items as a choice between levels. For QAGS, the criterion is the question its workers answered, asked of the whole summary: ``Is the summary factually supported by the article?''.

\begin{table}[!htb]
\centering\footnotesize
\setlength{\tabcolsep}{4pt}
\caption{Evaluation benchmarks.}
\label{tab:bench}
\begin{tabular}{@{}l>{\raggedright\arraybackslash}p{6.0cm}lr@{}}
\toprule
Benchmark & What is judged & Labels from & Judgments \\
\midrule
HealthBench \citep{arora2025healthbench} & health conversations, consensus criteria (met / not met) & 2--5 physicians & 29,511 \\
BiGGen \citep{kim2025biggen} & responses with instance-specific rubrics, nine capabilities (5 levels) & human raters & 3,194 \\
ELLIPSE \citep{crossley2023ellipse} & learner essays, overall proficiency and six traits (5 levels) & two trained raters & 45,374 \\
SummEval \citep{fabbri2021summeval} & news summaries, four dimensions (5 levels) & three experts & 6,400 \\
JudgmentBench, binary \citep{judgmentbench} & legal answers, checklist items (met / not met) & practising lawyers & 17,365 \\
JudgmentBench, multi-level \citep{judgmentbench} & legal answers, counted errors (4 levels) & practising lawyers & 2,782 \\
QAGS-CNN \citep{wang2020qags} & multi-sentence model summaries of CNN/DailyMail articles (supported / not) & three crowd workers & 235 \\
QAGS-XSUM \citep{wang2020qags} & one-sentence model summaries of BBC articles (supported / not) & three crowd workers & 239 \\
\bottomrule
\end{tabular}
\end{table}

\subsection{Baselines}
\textbf{Jev} \citep{jev} receives one \emph{score} question per judgment: the rubric's criterion is the question's \texttt{instructions}, and its level descriptions, lowest first, are the question's \texttt{criteria}. Jev returns a probability for each criterion, which we use as its distribution over levels. \textbf{Qwen3-8B} \citep{qwen3}, the same base model as RubriX's encoder, is used as a zero-shot judge with thinking disabled: the levels are listed lowest first and lettered A, B, C, \ldots, and the level distribution is the softmax of the first generated token's log-probabilities restricted to those letters. \textbf{G-Eval} \citep{liu2023geval} results are copied from its paper for the two benchmarks it shares with us, SummEval (its Table~1) and QAGS (its Table~3), with GPT-4 (G-Eval-4) and GPT-3.5 (G-Eval-3.5).

\subsection{Metrics}
\textbf{Exact accuracy} is the share of judgments whose most likely level equals the gold level, defined as the level nearest to the raters' mean score, with exact halves assigned to the lower level (on the held-out set, the dataset's own label). For QAGS, whose human score is the share of a summary's sentences that the workers' majority judged supported, a summary is labelled supported only if every one of its sentences is, and unsupported if even one is not. \textbf{Within-one accuracy} counts a prediction at most one level away as correct; it is reported for multi-level benchmarks only, since on two levels it is always 100\%. Every judge is scored on the same judgments.

\textbf{Ranking correlation}: we report summary-level Spearman $\rho$ and Kendall $\tau$ \citep{bhandari2020reeval}. Judgments are grouped by input and criterion, for example one SummEval article's 16 summaries for one dimension; within each group the judge's scores are correlated with the human scores (Kendall $\tau$ as $\tau$-b, which handles ties), so the question is whether a judge orders the responses to the same input the way people do. Group correlations are averaged within each criterion, then within each category, and the reported AVG is the mean over a benchmark's categories. A judge's score is its probability-weighted level, $\sum_k p_k\,k$, the same score G-Eval uses (Jev's probabilities, returned to two decimals, are renormalised to sum to one). Groups with fewer than three responses, or where the human score or the judge's score never changes, are skipped. QAGS has one summary per article, so, as in G-Eval, each subset is one correlation over all its summaries.

\subsection{Implementation details}
The encoder is Qwen3-8B in bf16 on one H100 80GB GPU, with gradient checkpointing. LoRA uses rank 16, $\alpha=32$ and dropout 0.05 on all seven attention and MLP projections. We train with AdamW (no weight decay; learning rates $10^{-4}$ for the adapters and $5\times10^{-4}$ for the heads), micro-batches of 8 with 4 accumulation steps (an effective batch of 32), 5\% linear warm-up then linear decay, and gradient clipping at 1.0, for one epoch of 1,071 steps. The checkpoint with the lowest log-loss on a fixed sample of the validation set was kept (step 1,068). Training took 10,212 seconds (2.8 hours) over 32.4M tokens. For evaluation, the adapters are merged into the encoder, which is served in bf16 with vLLM \citep{kwon2023vllm}.

\section{Results}

\subsection{Held-out rubrics}
\begin{figure}[!htb]
\centering
\includegraphics[width=0.72\linewidth]{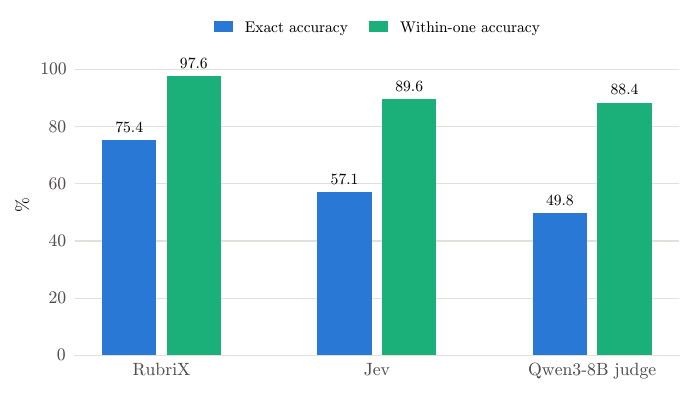}
\caption{Held-out test set: 6,749 judgments on rubrics whose criteria and level descriptions never appear in training.}
\label{fig:heldout}
\end{figure}

The test set held out from the training mix contains rubrics whose criteria and level descriptions never appear in training: 6,749 judgments, from the Feedback Collection (4,999), the proprietary dataset (1,172), HelpSteer3 principles (378), HelpSteer (100) and HelpSteer2 (100). RubriX is far ahead (Figure~\ref{fig:heldout}): 75.4\% exact accuracy against 57.1\% for Jev and 49.8\% for the Qwen3-8B judge, and 97.6\% within one level against 89.6\% and 88.4\%. These rubrics come from the same sources as the training data, so this result measures generalisation to new criteria of a familiar kind.

\subsection{Human-labelled benchmarks: accuracy}
\begin{figure}[!htb]
\centering
\includegraphics[width=\linewidth]{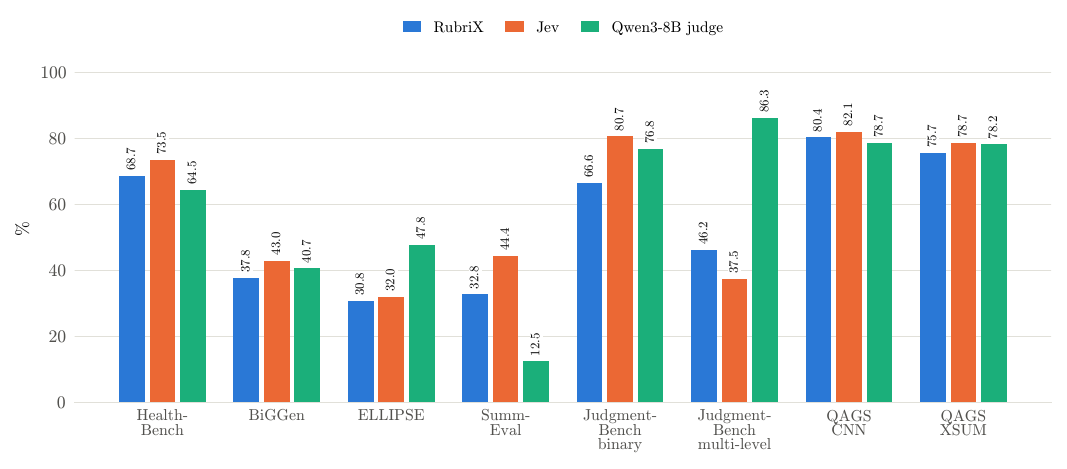}
\caption{Exact accuracy on the human-labelled benchmarks.}
\label{fig:bench}
\end{figure}

\begin{figure}[!htb]
\centering
\includegraphics[width=0.66\linewidth]{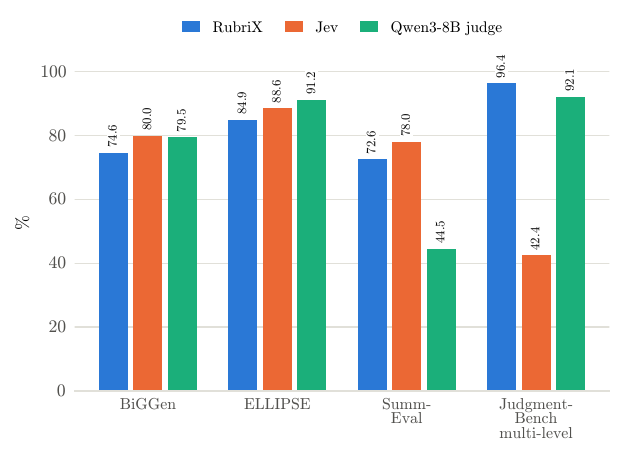}
\caption{Within-one accuracy on the multi-level benchmarks.}
\label{fig:bench-w1}
\end{figure}

Figures~\ref{fig:bench} and~\ref{fig:bench-w1} report accuracy on the benchmarks. Jev has the highest exact accuracy on six of the eight sets (HealthBench, BiGGen, SummEval, JudgmentBench binary and both QAGS subsets), but RubriX is close behind on most of them: by 1.7 points on QAGS-CNN, 3.0 on QAGS-XSUM, 4.8 on HealthBench, 5.2 on BiGGen and 1.2 on ELLIPSE, where the Qwen3-8B judge leads. The larger gaps are SummEval (11.6 points) and JudgmentBench binary (14.1 points). The Qwen3-8B judge is the most accurate on ELLIPSE and JudgmentBench multi-level, where RubriX is ahead of Jev. RubriX is close to Jev on QAGS (80.4\% against 82.1\% on CNN, 75.7\% against 78.7\% on XSUM) and has the best within-one accuracy on JudgmentBench multi-level (96.4\%), where Jev reaches 42.4\%.

\subsection{Human-labelled benchmarks: ranking correlation}
These correlations answer a different question from accuracy: given several responses to the same input, does the judge put them in the same order as people do, even when it does not land on the exact level? Table~\ref{tab:corr} reports Spearman $\rho$ and Kendall $\tau$ per category. Jev has the highest average correlation on five of the six benchmarks; on BiGGen the Qwen3-8B judge is level with it (0.527 against 0.526 in $\rho$).

Against G-Eval, on the two benchmarks it reports:
\begin{itemize}
  \item \emph{SummEval}: RubriX averages $\rho=0.483$, above G-Eval-3.5 (0.401) and below G-Eval-4 (0.514); on coherence it matches G-Eval-4 (0.581 against 0.582). Jev averages 0.541, above G-Eval-4, and is best on three of the four dimensions; G-Eval-4 is best on relevance.
  \item \emph{QAGS}: RubriX has the highest Spearman correlation on QAGS-CNN (0.719, against 0.715 for Jev and 0.685 for G-Eval-4), but is lower on QAGS-XSUM (0.521, against 0.537 for G-Eval-4 and 0.653 for Jev), whose one-sentence summaries are more abstractive. Averaged over both subsets, RubriX (0.620) is level with G-Eval-4 (0.611) and the Qwen3-8B judge (0.616), and Jev (0.684) is ahead.
\end{itemize}
RubriX is above G-Eval-3.5 in every SummEval dimension and both QAGS subsets, in both $\rho$ and $\tau$.

On ELLIPSE and HealthBench, RubriX ranks second, between Jev and the Qwen3-8B judge. On BiGGen, where each group holds only about four responses to one instance, it is third. On JudgmentBench, every judge is close to zero ($\rho \le 0.068$ on average): ordering the answers to the same legal task by their checklist and error items is not something any of these judges does reliably.

\begin{table}[!htb]
\centering\scriptsize
\setlength{\tabcolsep}{3.2pt}
\caption{Ranking correlation with human scores: summary-level Spearman $\rho$ and Kendall $\tau$ within groups of responses to the same input, averaged per category; AVG is the mean over a benchmark's categories. G-Eval values are copied from its paper (Table~1 for SummEval, Table~3 for QAGS); it does not report the other benchmarks. Bold marks the best judge in each row and underline the second best, separately for $\rho$ and $\tau$. A group is one article's summaries for one dimension (SummEval), the essays written to one prompt for one trait (ELLIPSE), one conversation's replies for one criterion (HealthBench), one instance's responses (BiGGen) or the answers to one task for one rubric item (JudgmentBench); for QAGS each subset is one group.}
\label{tab:corr}
\begin{tabular}{@{}llcccccccccc@{}}
\toprule
& & \multicolumn{2}{c}{RubriX} & \multicolumn{2}{c}{Jev} & \multicolumn{2}{c}{Qwen3-8B judge} & \multicolumn{2}{c}{G-Eval-4} & \multicolumn{2}{c}{G-Eval-3.5} \\
\cmidrule(lr){3-4} \cmidrule(lr){5-6} \cmidrule(lr){7-8} \cmidrule(lr){9-10} \cmidrule(lr){11-12}
Benchmark & Category & $\rho$ & $\tau$ & $\rho$ & $\tau$ & $\rho$ & $\tau$ & $\rho$ & $\tau$ & $\rho$ & $\tau$ \\
\midrule
SummEval & Coherence & 0.581 & 0.450 & \textbf{0.649} & \textbf{0.510} & 0.432 & 0.330 & \underline{0.582} & \underline{0.457} & 0.440 & 0.335 \\
 & Consistency & 0.496 & 0.413 & \textbf{0.559} & \textbf{0.473} & 0.458 & 0.377 & \underline{0.507} & \underline{0.425} & 0.386 & 0.318 \\
 & Fluency & 0.435 & 0.360 & \textbf{0.484} & \textbf{0.402} & \underline{0.461} & \underline{0.381} & 0.455 & 0.378 & 0.424 & 0.347 \\
 & Relevance & 0.422 & 0.328 & \underline{0.474} & \underline{0.370} & 0.360 & 0.280 & \textbf{0.547} & \textbf{0.433} & 0.385 & 0.293 \\
 & \textit{AVG} & 0.483 & 0.388 & \textbf{0.541} & \textbf{0.439} & 0.428 & 0.342 & \underline{0.514} & \underline{0.418} & 0.401 & 0.320 \\
\midrule
QAGS & QAGS-CNN & \textbf{0.719} & \underline{0.583} & \underline{0.715} & 0.579 & 0.656 & 0.518 & 0.685 & \textbf{0.591} & 0.516 & 0.410 \\
 & QAGS-XSUM & 0.521 & 0.427 & \textbf{0.653} & \textbf{0.547} & \underline{0.576} & \underline{0.472} & 0.537 & \underline{0.472} & 0.406 & 0.343 \\
 & \textit{AVG} & \underline{0.620} & 0.505 & \textbf{0.684} & \textbf{0.563} & 0.616 & 0.495 & 0.611 & \underline{0.525} & 0.461 & 0.377 \\
\midrule
ELLIPSE & Overall & \underline{0.589} & \underline{0.463} & \textbf{0.651} & \textbf{0.520} & 0.520 & 0.403 & -- & -- & -- & -- \\
 & Cohesion & \underline{0.510} & \underline{0.395} & \textbf{0.579} & \textbf{0.453} & 0.475 & 0.364 & -- & -- & -- & -- \\
 & Syntax & \underline{0.516} & \underline{0.399} & \textbf{0.590} & \textbf{0.462} & 0.494 & 0.379 & -- & -- & -- & -- \\
 & Vocabulary & 0.516 & 0.402 & \textbf{0.569} & \textbf{0.447} & \underline{0.520} & \underline{0.406} & -- & -- & -- & -- \\
 & Phraseology & \underline{0.472} & \underline{0.362} & \textbf{0.566} & \textbf{0.441} & 0.467 & 0.358 & -- & -- & -- & -- \\
 & Grammar & \underline{0.532} & \underline{0.410} & \textbf{0.615} & \textbf{0.482} & 0.424 & 0.324 & -- & -- & -- & -- \\
 & Conventions & 0.470 & 0.360 & \textbf{0.636} & \textbf{0.500} & \underline{0.497} & \underline{0.380} & -- & -- & -- & -- \\
 & \textit{AVG} & \underline{0.515} & \underline{0.399} & \textbf{0.601} & \textbf{0.472} & 0.485 & 0.373 & -- & -- & -- & -- \\
\midrule
HealthBench & Communication & \underline{0.286} & \underline{0.259} & \textbf{0.361} & \textbf{0.332} & 0.226 & 0.204 & -- & -- & -- & -- \\
 & Complex responses & 0.131 & 0.120 & \textbf{0.162} & \textbf{0.149} & \underline{0.155} & \underline{0.139} & -- & -- & -- & -- \\
 & Context seeking & \underline{0.484} & \underline{0.437} & \textbf{0.565} & \textbf{0.518} & 0.317 & 0.285 & -- & -- & -- & -- \\
 & Emergency referrals & \underline{0.429} & \underline{0.388} & \textbf{0.493} & \textbf{0.453} & 0.357 & 0.322 & -- & -- & -- & -- \\
 & Global health & \textbf{0.438} & \textbf{0.398} & \underline{0.436} & \textbf{0.398} & 0.340 & \underline{0.306} & -- & -- & -- & -- \\
 & Health data tasks & 0.314 & 0.286 & \textbf{0.584} & \textbf{0.535} & \underline{0.355} & \underline{0.323} & -- & -- & -- & -- \\
 & Hedging & 0.154 & 0.140 & \textbf{0.483} & \textbf{0.447} & \underline{0.160} & \underline{0.147} & -- & -- & -- & -- \\
 & \textit{AVG} & \underline{0.320} & \underline{0.290} & \textbf{0.441} & \textbf{0.405} & 0.273 & 0.246 & -- & -- & -- & -- \\
\midrule
BiGGen & Grounding & 0.549 & 0.513 & \textbf{0.643} & \textbf{0.598} & \underline{0.632} & \underline{0.577} & -- & -- & -- & -- \\
 & Instruction following & 0.405 & 0.359 & \textbf{0.560} & \textbf{0.507} & \underline{0.550} & \underline{0.492} & -- & -- & -- & -- \\
 & Multilingual & 0.418 & 0.342 & \underline{0.427} & \underline{0.365} & \textbf{0.474} & \textbf{0.408} & -- & -- & -- & -- \\
 & Planning & 0.507 & 0.444 & \underline{0.539} & \underline{0.489} & \textbf{0.580} & \textbf{0.512} & -- & -- & -- & -- \\
 & Reasoning & 0.453 & 0.408 & \underline{0.533} & \underline{0.496} & \textbf{0.542} & \textbf{0.498} & -- & -- & -- & -- \\
 & Refinement & 0.363 & 0.320 & \textbf{0.577} & \textbf{0.513} & \underline{0.514} & \underline{0.468} & -- & -- & -- & -- \\
 & Safety & 0.362 & 0.325 & \underline{0.527} & \underline{0.476} & \textbf{0.545} & \textbf{0.478} & -- & -- & -- & -- \\
 & Theory of mind & 0.404 & 0.362 & \textbf{0.458} & \textbf{0.415} & \underline{0.438} & \underline{0.390} & -- & -- & -- & -- \\
 & Tool usage & 0.442 & 0.394 & \underline{0.465} & \underline{0.414} & \textbf{0.466} & \textbf{0.424} & -- & -- & -- & -- \\
 & \textit{AVG} & 0.434 & 0.385 & \underline{0.526} & \textbf{0.475} & \textbf{0.527} & \underline{0.472} & -- & -- & -- & -- \\
\midrule
JudgmentBench & Substance & \underline{0.082} & \underline{0.067} & \textbf{0.114} & \textbf{0.099} & 0.073 & 0.059 & -- & -- & -- & -- \\
 & Structure & \underline{$-$0.002} & \underline{$-$0.001} & \textbf{0.098} & \textbf{0.091} & $-$0.037 & $-$0.030 & -- & -- & -- & -- \\
 & Style & \textbf{0.047} & \textbf{0.037} & \underline{0.010} & \underline{0.001} & $-$0.030 & $-$0.025 & -- & -- & -- & -- \\
 & Negative points & \textbf{0.051} & \textbf{0.041} & \underline{0.048} & \underline{0.039} & 0.003 & 0.003 & -- & -- & -- & -- \\
 & \textit{AVG} & \underline{0.045} & \underline{0.036} & \textbf{0.068} & \textbf{0.058} & 0.002 & 0.002 & -- & -- & -- & -- \\
\bottomrule
\end{tabular}

\end{table}

\subsection{Latency and rubric size}
\label{sec:latency}
A judge that reads the rubric inside its input pays for every level on every judgment. RubriX embeds each level description once per rubric, so a judgment on a rubric it has seen needs only the input embedded. We measured latency one request at a time, on dummy text so that token counts are exact: an 800-token response, a 30-token criterion, and level texts of a fixed length, with 110 judgments per setting (30 for the cache miss at 50 and 100 levels, whose rubrics hold 75,000 and 150,000 tokens) and a different response for every judgment. Two settings were varied: the number of levels, from 2 to 100 levels of 1,500 tokens each, and the length of each level, from 10 to 3,000 tokens with 5 levels. RubriX (the evaluated model's weights, adapters merged) and the Qwen3-8B judge ran on one H100 80GB with vLLM, with vLLM's prefix cache off for both; Jev was called through its API.

\begin{figure}[!htb]
\centering
\includegraphics[width=\linewidth]{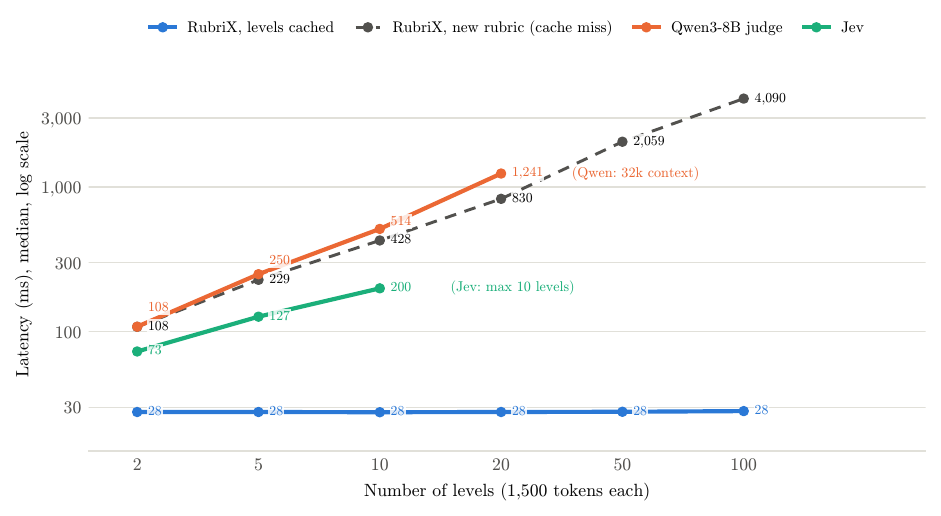}
\caption{Median latency per judgment, one request at a time, against the number of levels (1,500 tokens each); labels give the median in ms. Latency axis on a log scale. Jev's line has an approximate network time subtracted: 232.6\,ms, the median time, over 50 fresh connections from our client, from the end of the connection handshake to the first byte of the reply to an empty request to its API. RubriX and the Qwen3-8B judge were timed on the H100 itself and show processing only. Jev accepts at most 10 levels and the Qwen3-8B judge's 32k context holds at most 20 levels of this length.}
\label{fig:latency}
\end{figure}

\begin{figure}[!htb]
\centering
\includegraphics[width=\linewidth]{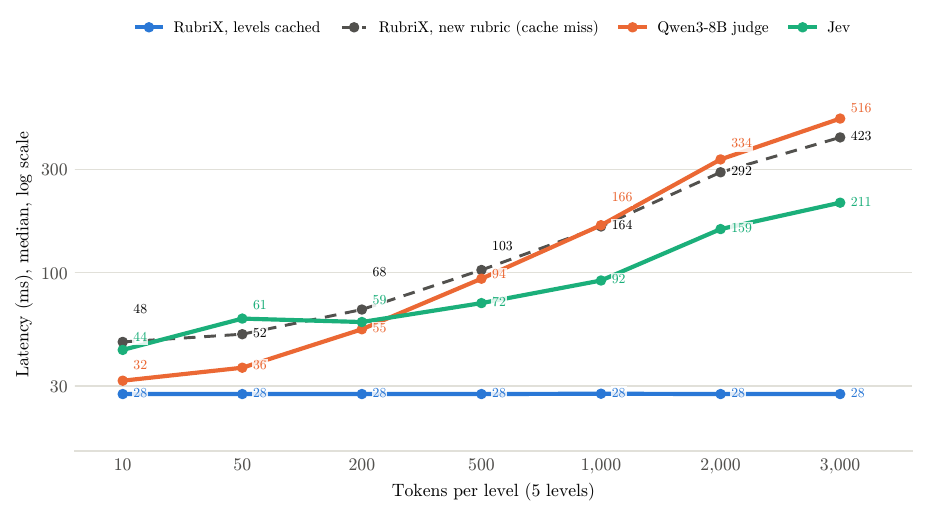}
\caption{Median latency per judgment against the length of each level description (5 levels), as in Figure~\ref{fig:latency}.}
\label{fig:latency-len}
\end{figure}

Figures~\ref{fig:latency} and~\ref{fig:latency-len} show the result. With its level embeddings cached, RubriX takes 27.5--28.3\,ms per judgment in every setting, up to a 100-level rubric of 150,000 tokens. The Qwen3-8B judge grows from 108.5\,ms at 2 levels to 1,241.3\,ms at 20 levels of 1,500 tokens, the most its 32k context can hold, and from 31.7\,ms to 516.2\,ms as levels grow from 10 to 3,000 tokens. Jev grows from 73.1\,ms to 199.7\,ms between 2 and 10 levels, the maximum its score questions accept, and from 44.0\,ms to 210.9\,ms with level length. A cache miss, when RubriX meets a new rubric and embeds its levels in the same request, costs about as much as the Qwen3-8B judge: 108.2\,ms at 2 levels, 4,090.2\,ms for a 100-level, 150,000-token rubric, and 422.6\,ms at 5 levels of 3,000 tokens. It happens once per rubric.

\paragraph{Training cost.} RubriX trained 62.5M parameters (43.6M adapter and 18.9M head parameters) for one epoch in 2.8 hours on one H100. At inference, each judgment needs one encoder forward pass for its input (the conversation, the response and the criterion). Each level description is embedded in its own pass only once per rubric, and the vectors are reused for every response judged on that rubric; no text is generated.

\section{Limitations}
\begin{itemize}
  \item \textbf{Separate encoding.} The input and the level descriptions are encoded separately and meet only at the final similarity, so the model never reads a response alongside a level description. This limits verdicts that depend on checking a specific fact against the rubric.
  \item \textbf{Model capacity.} RubriX uses an 8B encoder with small projection heads (18.9M parameters), which likely limits how far it generalises beyond the rubric styles it was trained on; we have not trained larger encoders to test this. Continuing training for a second epoch, from the final model on a partly refreshed data mix, raised validation log-loss at every check, so training stopped early and the first-epoch model was kept: with this encoder and head size, more training on this data does not help.
  \item \textbf{Dependence on fine-tuning data.} RubriX responds strongly to fine-tuning: on held-out rubrics from its training sources it leads Jev by 18.3 percentage points in exact accuracy, but on human-labelled benchmarks of other formats it trails Jev on most sets. It is therefore most useful when it can be fine-tuned on rubrics like the ones it will be applied to; on unfamiliar rubric styles, without such data, a general-purpose judge such as Jev is stronger.
\end{itemize}

\section{Future Work}
\begin{itemize}
  \item \emph{Larger encoders.} Training RubriX on bigger base models, to measure how much of the gap to Jev on unfamiliar rubric styles is a matter of model size.
  \item \emph{More training data.} Scaling the training data, in size and in the variety of criteria and formats, and analysing how accuracy and ranking on held-out rubrics and human-labelled benchmarks change with it.
  \item \emph{Domain-specific fine-tuning.} Fine-tuning on data from a particular use case and comparing how the model then performs on that domain's rubrics against general rubrics.
  \item \emph{Rubrics with hundreds of levels.} Fine-tuning for scoring tasks that need very many levels. RubriX's cost does not grow with the number of levels, while an LLM judge must fit every level description in its context, which at this scale needs long-context models, for example with extended rotary position embeddings (RoPE) \citep{su2024roformer}; we will compare the two in accuracy and efficiency.
\end{itemize}

\section{Conclusion}
RubriX, whose architecture is inspired by contrastive language models, scores rubrics with a LoRA-adapted encoder, returning a probability for every level without generation. On held-out rubrics it is well ahead of Jev and a Qwen3-8B judge. On human-labelled benchmarks it is competitive: it is within 6 percentage points of Jev's accuracy on most sets, Jev keeps a larger lead on SummEval and JudgmentBench binary and ranks best on average, while RubriX leads on QAGS-CNN and matches G-Eval with GPT-4 on SummEval coherence. Its clearest advantage is cost: with cached level embeddings its latency does not depend on the size of the rubric, so long, many-level rubrics cost no more to apply than short ones. RubriX is therefore best suited to settings where a few labelled examples from the target domain are available to fine-tune on: trained with at most 25 examples per criterion, it beats Jev and the Qwen3-8B judge by a wide margin on held-out rubrics from the same sources (75.4\% exact accuracy against 57.1\% and 49.8\%), its fine-tuning took 2.8 hours on a single H100, and each judgment then costs one encoder pass.

\setlength{\bibsep}{3pt}

\end{document}